\documentclass[sigconf,nonacm]{acmart}

\usepackage{amsmath,amssymb,amsfonts}
\usepackage{algorithmic}
\usepackage{graphicx}
\usepackage{textcomp}
\usepackage{xcolor}

\usepackage{algorithm}
\usepackage{xspace}
\usepackage{array}
\usepackage{multirow}
\usepackage{tikz}
\usepackage{etoolbox}
\usepackage{makecell}
\usepackage[caption=false]{subfig}
\usepackage{bm}
\usepackage[T1]{fontenc}

\DeclareMathOperator*{\argmax}{arg\,max}

\newcommand{\fig}[1]{Figure~\ref{#1}}
\newcommand{\sect}[1]{Section~\ref{#1}}
\newcommand{\tab}[1]{Table~\ref{#1}}
\newcommand{\algo}[1]{Algorithm~\ref{#1}}
\newcommand{\circlednum}[1]{\tikz[baseline=-0.7ex]{\node[draw, circle, fill=black, text=white, inner sep=0.05mm]{#1};}\ignorespacesafterend}
\robustify{\circlednum}
\newcommand{\proposed}{Agent-X\xspace}
\newcommand{\prefill}{PromptWeaver\xspace}
\newcommand{\decode}{ExSpec\xspace}

\title{\proposed: Full Pipeline Acceleration of On-device AI Agents}

\author{Jinha Chung}
\affiliation{
  \institution{KAIST}
  \country{Republic of Korea}
  \city{Daejeon}
}
\email{jinha.chung@kaist.ac.kr}

\author{Byeongjun Shin}
\affiliation{
  \institution{KAIST}
  \country{Republic of Korea}
  \city{Daejeon}
}
\email{byeongjun.shin@kaist.ac.kr}

\author{Jiin Kim}
\affiliation{
  \institution{KAIST}
  \country{Republic of Korea}
  \city{Daejeon}
}
\email{jiin.kim@kaist.ac.kr}

\author{Minsoo Rhu}
\affiliation{
  \institution{KAIST}
  \country{Republic of Korea}
  \city{Daejeon}
}
\email{mrhu@kaist.ac.kr}

\begin{document}

\begin{abstract}
LLM-based agents deliver state-of-the-art performance across tasks but incur high end-to-end latency on edge devices. We introduce \proposed, a software-only, accuracy-preserving framework that accelerates both the prefill and decode stages of on-device agent workloads. \proposed's two key components rewrite prompts to leverage prefix caching tailored to agent-specific input-token patterns and enable LLM-free speculative decoding for fast token generation with minimal overhead. On representative agentic workloads, \proposed achieves a 1.61$\times$ end-to-end speedup in real systems with no accuracy loss and can be seamlessly integrated into existing on-device AI agents. To the best of our knowledge, ours is the first to systematically characterize and eliminate latency bottlenecks in on-device agents.
\end{abstract}

\maketitle
\section{Introduction} \label{sect:introduction}

\begin{figure}[t!] \centering
\includegraphics[width=\columnwidth]{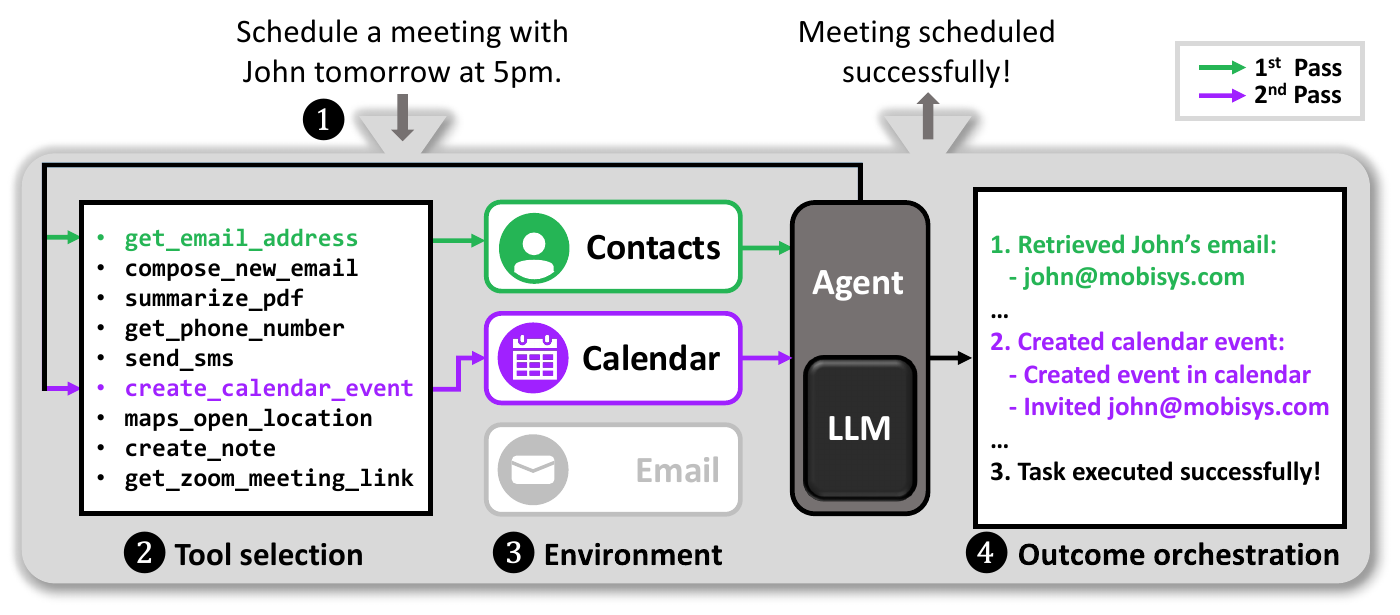}
\caption{
Overview of an agentic system.
}
\Description{Overview of an agentic system.}
\label{fig:agent_overview}
\end{figure}

The ``ChatGPT effect'' has taken the world by storm, and Large Language Models (LLMs) are now embedded in various applications that drive our daily lives. 
LLM-based AI agents elevate the applicability of LLMs through ``tool calling.'' Equipped with external tools, an LLM can \textit{interact} with its environment and autonomously execute tasks from start to finish, without further user intervention. As shown in \fig{fig:agent_overview}, given a user query (\circlednum{1}), the AI agent selects the appropriate tools to handle the request (\circlednum{2}). The agent then interacts with the environment (e.g., ``Contacts'', ``Calendar'', and ``Email'') by calling the selected tool (\circlednum{3}) and reflecting on prior tool output(s) to select the next action (\circlednum{4}). In academia, a rich body of prior work~\cite{react,reflexion,llmcompiler,lats} has improved agent task accuracy, while the industry has already started deploying LLM-based agents~\cite{deep_research,anthropic_agent,gemini_cli,chatgpt_agent,manus}, boosting user productivity and convenience.

On-device AI agents, which run entirely on a user’s local device, provide two unique benefits over cloud-based agents: availability and privacy. On-device agents are always available to the user for immediate use, regardless of the user's situation (lack of internet access) or the cloud provider's situation (server outage). Furthermore, growing concerns over data misuse call for measures to guarantee privacy in using LLMs, making on-device agents an attractive solution.

Building on this trend, various hardware and software have been introduced to ease the development and deployment of LLMs at the edge~\cite{copilot_pc,qualcomm_snapdragon_x_elite,amd_ryzen_ai_max,apple_m4_pro_max,qualcomm_snapdragon_hexagon_sdk,apple_foundation_models_framework}. Despite these advances, on-device AI agents still suffer from suboptimal latency, even for simple tasks, due to the resource-constrained nature of edge computing. Unlike cloud-based LLMs whose primary performance bottleneck lies in the decode stage, this paper makes the key observation that on-device agents spend a significant amount of time in \emph{both} the prefill and decode stages.  This key insight underscores the need for full-system acceleration techniques that address all key components of the agentic system pipeline on edge hardware.

\begin{figure*}[t!] \centering
\includegraphics[width=1.0\textwidth]{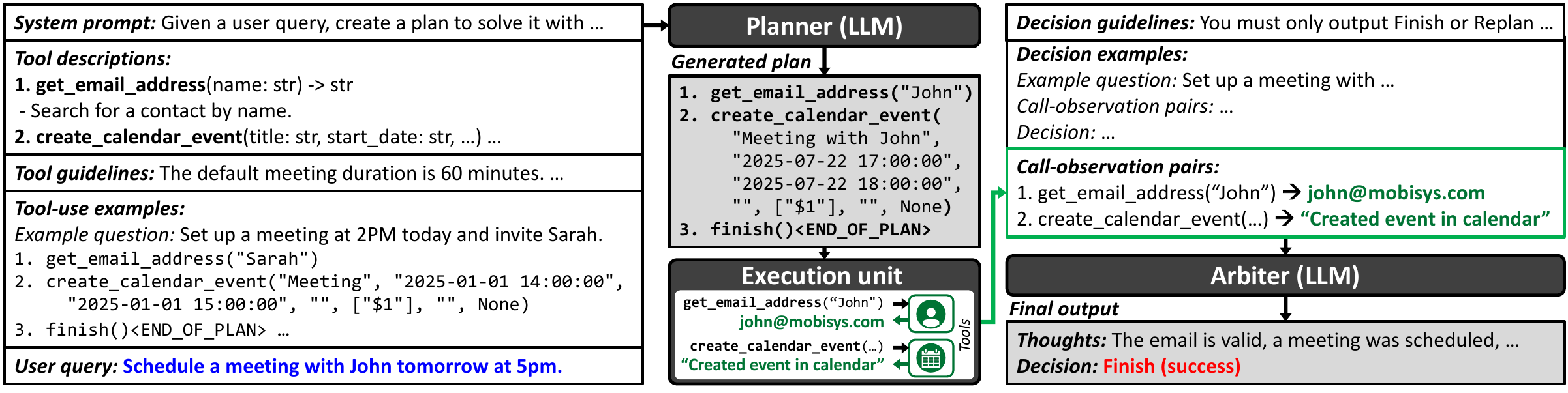}
\caption{
Structure of plan-out agents. The whole pipeline consists of two LLMs (Planner and Arbiter) and a series of tool calls (carried out by Execution unit) to fully serve the user query (blue).
\Description{Structure of plan-out agents.}
}
\label{fig:plan-out_agents}
\end{figure*}

To this end, we propose \proposed, a purely software end-to-end acceleration scheme for on-device agents that does not degrade accuracy. To the best of our knowledge, this is the first work to provide a detailed system-level characterization of on-device AI agents. Building on this analysis, we introduce a full-pipeline acceleration solution that exploits both the algorithmic traits of agents and the hardware characteristics of edge environments, as detailed below.

{\bf On-device agents analysis.} We analyze the execution of LLM-based agents and identify two LLM instances as primary bottlenecks. Our analysis shows that, unlike conventional server-based, conversation-oriented applications where latency is dominated by decoding, both the prefill and decode stages contribute significantly to end-to-end latency in on-device environments due to the agentic workflow and hardware constraints. We further characterize these stages at the \textit{token level} and observe two key properties. First, during prefill, the prompt structure limits the applicability of optimizations such as prefix caching. Second, during decode, the output is largely grounded in few-shot examples and does not fully exploit the LLM’s reasoning capability.

{\bf Accuracy-preserving acceleration algorithm.} Building on our characterization, we propose \emph{\proposed}, which combines \emph{\prefill} and \emph{\decode} to accelerate the prefill and decode stages of on-device agentic LLMs, respectively. \prefill dynamically reconstructs the input prompt to enable efficient prefix caching, reducing online computation and substantially accelerating the prefill stage. We also note that, while speculative decoding~\cite{specdec, sps} is widely used to speed up LLM decoding, its reliance on additional LLMs hinders deployment on edge devices. To address this, \decode introduces a lightweight, prompt-aware draft model that enables efficient speculative decoding at the edge.

{\bf Real system integration.} We implement \proposed with Apple's MLX-LM~\cite{mlx_lm} and MLX-engine~\cite{mlx_engine}, and integrate it with TinyAgent~\cite{tinyagent} for a full system acceleration of on-device agents. Our evaluation shows that \prefill and \decode speed up the prefill and decode stages of on-device agentic LLMs by $1.97\times$ and $1.73\times$, respectively, achieving an average end-to-end task latency improvement of $1.61\times$.

Overall, \proposed accelerates on-device agents with minimal resource overhead. Its lightweight, accuracy-preserving, and purely software design allows direct integration into existing on-device agentic workflows, delivering immediate speedups and enabling fast, private agents at the edge.

\section{Background} \label{sect:background}

\subsection{LLM-based Agents} \label{sect:agents}
{\bf Agentic workflows.} LLM-based agents enhance LLMs' capabilities by interacting with the external environment. The means of interaction are referred to as \emph{tools}, which are APIs available within the system, ranging from built-in system functions (e.g., access to file systems) to third-party application APIs (e.g., sending emails). For the LLM to be aware of such tools, the list of available tools, as well as their \textit{descriptions} (how to use them), \textit{guidelines} (caveats), and \textit{tool-use examples} (few-shot examples~\cite{gpt3}) are conveyed through the input prompt~\cite{mcp, rag_mcp, toolformer, hugginggpt, llama3} (\fig{fig:plan-out_agents}). The outputs of the tool calls, called \textit{observations}, are returned to the LLM so it can decide whether to retry or continue. This fundamental workflow, introduced in ReAct~\cite{react}, has been extended with new mechanisms, such as adding reflection capabilities~\cite{reflexion} or optimal-path search~\cite{lats} to improve task accuracy.

Among existing approaches, agents that plan the full execution path \textit{before} tool calling are gaining traction~\cite{llmcompiler, plan_and_act, less_is_more, fused_parallel_function_calling}. Unlike ReAct, these ``plan-out'' agents consider interactions among tools when forming the plan. In ReAct, planning one tool call and observing its output require separate LLM calls, so an $N$-step plan needs 2$\cdot$$N$ LLM calls. In LLMCompiler~\cite{llmcompiler}, a representative, state-of-the-art plan-out agent, the full plan is generated in one LLM call and all observations are verified in the second LLM call, cutting the total to two LLM calls (an $N$ times reduction). The global view provided with a plan-out agent's full planning capability is known to also improve accuracy~\cite{llmcompiler, plan_and_act}, and is employed in state-of-the-art agentic services like Gemini~3.0~\cite{gemini_3.0}.

{\bf Structure of plan-out agents.} \fig{fig:plan-out_agents} outlines the overall workflow of plan-out agents. First, given the user query highlighted in blue, the agent retrieves appropriate tools (explained in depth in \sect{sect:on_device_agents}) and constructs the prompt for the first LLM, the \textit{Planner}. The prompt contains the system prompt, tool descriptions, guidelines, tool-use examples, and the user query. Planner outputs a list of tools plus their arguments, which may be literals (e.g., name ``John'') or references to prior results (e.g., feeding the output of plan \#1 as argument for \texttt{create\_calendar\_event} with \$1).

The execution unit executes the plans in order, parallelizing the execution of tools without dependencies. Each tool call and its output is recorded, forming a list of ``call-observation'' pairs (green box) that is passed to the second LLM, the \textit{Arbiter}. Based on the input consisting of guidelines, examples, and the call-observation pairs, the Arbiter decides whether the request is satisfied; if not, it signals a retry.

\subsection{On-device AI Agents} \label{sect:on_device_agents}

\begin{figure}[t!] \centering
\subfloat[Offline generation of tool-use example database]{\includegraphics[width=\columnwidth]{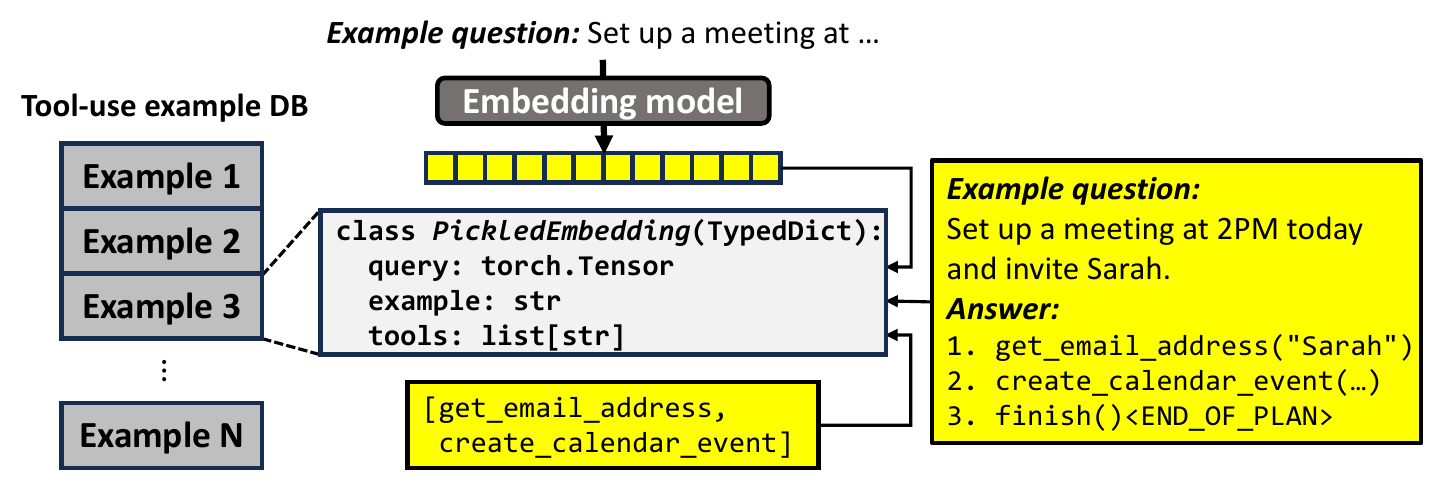}} \\
\subfloat[Tool retrieval]
{\includegraphics[width=\columnwidth]{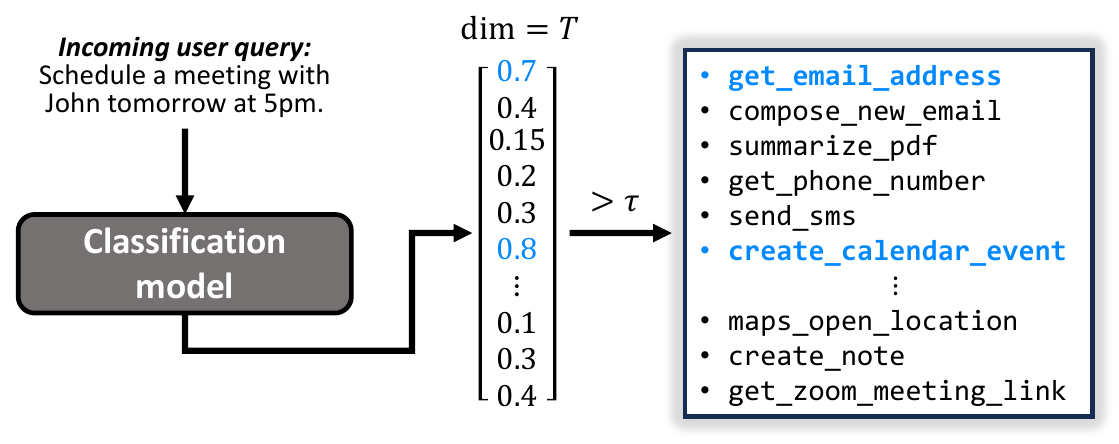}} \\
\subfloat[Tool-use example (few-shot example) retrieval]{\includegraphics[width=\columnwidth]{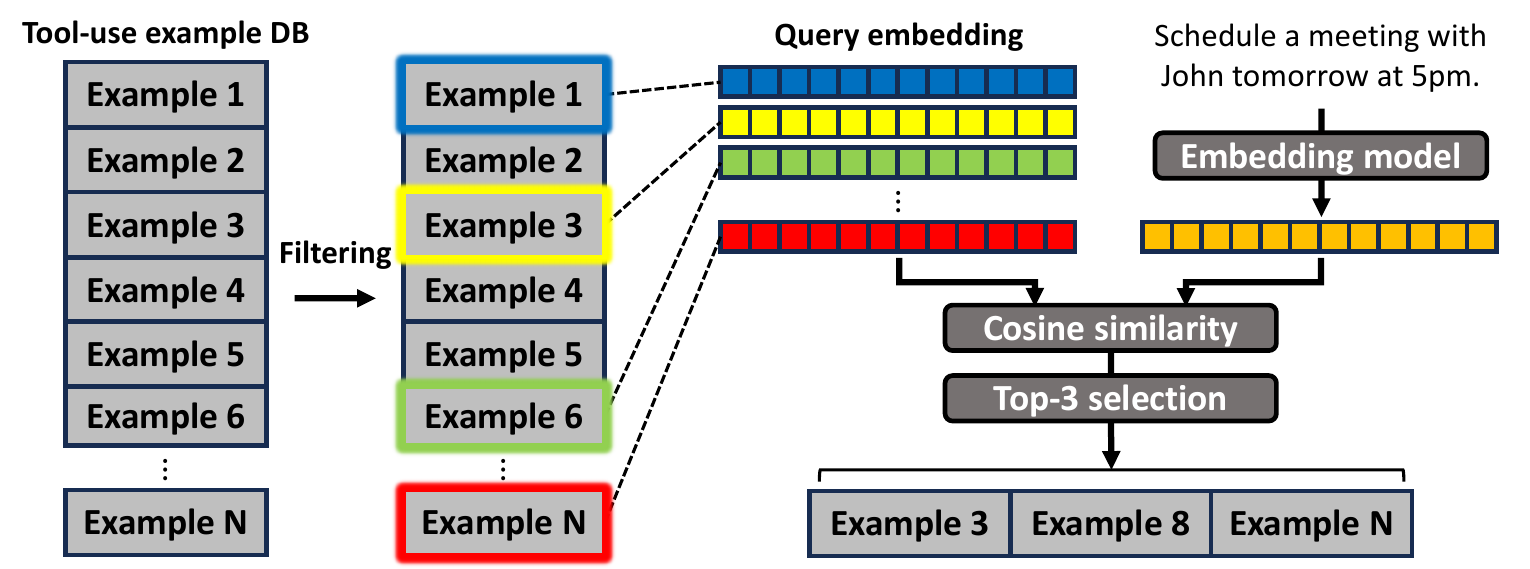}} \\
\caption{
Illustration of the ToolRAG process.
}
\Description{Illustration of the ToolRAG process.}
\label{fig:toolrag}
\end{figure}

{\bf On-device LLMs.} AI functionalities are now available on edge devices, appearing in various forms like voice transcription~\cite{apple_intelligence}, Circle-to-Search~\cite{circle_to_search}, and personal assistants~\cite{siri,bixby,gemini_assistant}. Among them, LLMs are increasingly being deployed at the edge, due to their powerful performance. To power edge workloads, compact LLMs for resource-constrained devices have emerged~\cite{gemini_nano,afm}. Hardware accelerators~\cite{cambricon_llm,edge_moe,edgemoe,edgellm,clone,decdec} and deployment schemes~\cite{llm_in_a_flash,edgelora,crosslm,xpert} for on-device LLMs have also been proposed, driving the integration of on-device LLMs into everyday life.

{\bf On-device agents.} On-device agents make use of on-device LLMs to power local agents executing entirely on the user’s device. This design mitigates the security and privacy risks inherent in cloud-based solutions. The local LLM processes user requests (e.g., setting reminders) by invoking OS or third-party APIs and completing tasks fully on-device. Among existing systems~\cite{siri, bixby, gemini_assistant}, TinyAgent~\cite{tinyagent} is an open-source macOS agent that fine-tunes LLMs for agentic tasks. It also introduces ToolRAG (Tool Retrieval Augmented Generation)~\cite{tinyagent_toolrag} for efficient tool and example selection.

{\bf Tool choice and tool-use examples.} ToolRAG comprises three components: (i) offline preparation of a tool-use example database, (ii) runtime tool retrieval, and (iii) tool-use example retrieval. A diverse set of user queries is collected offline. Each query is annotated with (1) a fixed-dimension text embedding, (2) the example plan to handle the query, and (3) the set of tools needed to complete the task. Each database entry is stored as a tuple of these three elements: (\texttt{query}, \texttt{example}, \texttt{tools}) in \fig{fig:toolrag}(a).

At runtime, the user query is fed into a lightweight classification model (\fig{fig:toolrag}(b)). This model outputs a probability score for each of the $T$ available tools. A threshold of $\tau$ is applied to these scores, retaining only tools whose probability exceeds $\tau$. This filtering ensures that only the most relevant tools (e.g., \texttt{get\_email\_address} and \texttt{create\_calendar\_event}) are considered. Next, the tool-use example database is filtered, leaving out entries that include tools not selected by the classification model. The cosine similarities between the user's query embedding and each embedding in the filtered tool-use example database are computed. The top-$K$ examples with the highest similarity scores are retrieved (\fig{fig:toolrag}(c)). Finally, the Planner prompt is assembled by combining: (1) detailed descriptions of the selected tools, (2) usage guidelines, and (3) the retrieved tool-use examples ($K=3$ in \fig{fig:plan-out_agents}). Overall, ToolRAG constructs a prompt with highly relevant examples, enabling the on-device agent's Planner to generate better execution plans to accomplish the request.

The mechanism of dynamic tool selection and contextual example retrieval in ToolRAG reflects a broadly adopted paradigm in agent design~\cite{mcp, rag_mcp, toolformer, hugginggpt, llama3}. Agent systems often use similar workflows to constrain their action spaces and guide generation with task-specific examples. For example, the Model Context Protocol (MCP)~\cite{mcp}, an emerging open-source standard for LLM tool access driven by Anthropic, OpenAI, and Google, specifies that tool descriptions and usage examples be provided to the model. Likewise, Google’s Function Calling guide for Gemini~\cite{google_dev_function_calling, hera_agent} recommends the filtering of available functions based on conversational context before issuing calls. These shared design principles demonstrate that dynamic tool selection and tool-use example retrieval is not unique to ToolRAG, but a widely employed design paradigm in designing agentic systems.

\subsection{LLM Inference Optimization Methods} \label{sect:llm_optimization}

\begin{figure}[t!] \centering
\includegraphics[width=\columnwidth]{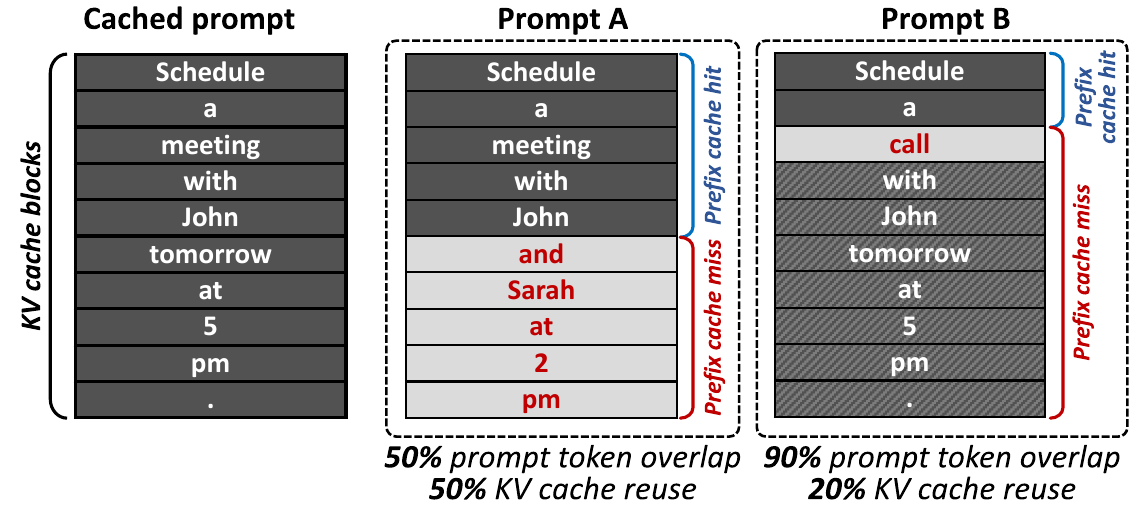} \\
\caption{
Illustration of applying prefix caching. Even though there exists substantial overlap between prompt B and the cached prompt, an early token mismatch limits the KV cache reuse.
}
\Description{Illustration of applying prefix caching.}
\label{fig:prefix_caching}
\end{figure}

LLM inference occurs in two stages: prefill and decode. During the prefill stage, the input prompt tokens are processed to populate the key-value (KV) cache, whose length grows with the number of tokens seen. In the decode stage, the model runs autoregressively, generating one token per step. Generally, prefill is compute-intensive, whereas decode is bounded by memory bandwidth~\cite{splitwise, pod_attention, sarathi, sarathi_serve, distserve}. Consequently, the latencies of conversational workloads running on cloud-based high-end GPU servers are known to be dominated by the decode stage (e.g., $>$ 95\% in \cite{wsc_llm, fdc, cent}).

{\bf Prefix caching.} In Transformer-based LLMs, each token attends only to preceding tokens. Therefore, when two inputs share a common prefix, they can reuse the same KV cache \textit{up to the first mismatched token.} \textit{Prefix caching}~\cite{vllm} exploits this by precomputing KV caches for shared prefixes, reducing prefill latency roughly proportional to the portion of cached tokens (i.e., amount of saved computation). As shown in \fig{fig:prefix_caching}, this technique works for prefixes of any length, but cache reuse halts at the first token mismatch, even if subsequent tokens are identical. Throughout this paper, \textit{cacheable tokens} refer to portions of prompts that prefix caching can be applied to with a single static prompt, and \textit{uncacheable tokens} refer to portions of prompts that cannot benefit from prefix caching due to an early token mismatch.

{\bf Speculative decoding.} An LLM's decode stage is memory bandwidth-bound because of its autoregressive, one-token-at-a-time nature, yielding low compute intensity. \textit{Speculative decoding}~\cite{specdec, sps} addresses this limitation by producing multiple \textit{draft tokens} per pass. \fig{fig:specdec} outlines the process, which uses a faster, less accurate \textit{draft model} to aid the generation of a slower, more accurate \textit{target model}.

First, the draft model autoregressively generates $N$ draft tokens (\circlednum{1}) with $N$ sequential forward passes. Next, the target model inputs the most recent token (\textit{``Schedule''}) concatenated with the $N$ draft tokens (\textit{``a'', ``meeting'', ``with'', ``Sarah''}) and enters the \textit{verification phase}. Here, for each draft token, the target model’s output logits and the draft model’s logits are compared, determining which tokens are accepted. The first mismatch (\textit{``Sarah''}) and all subsequent drafts are discarded (\circlednum{2}). Finally, the accepted draft tokens and the target model’s chosen next token (\textit{``John''}) form the input for the next speculation round (\circlednum{3}). Speculative decoding has been mathematically proven to yield outputs of comparable quality to standard autoregressive generation~\cite{specdec, sps}.

\section{Characterization and Motivation} \label{sect:motivation}

\subsection{Agentic Workload Characterization} \label{sect:workload_characterization}
\label{page:value_clarifications}
To understand the characteristics and implications of on-device agentic tasks, we measure and break down the end-to-end latency of TinyAgent~\cite{tinyagent}. We use 1,022 examples from the TinyAgent fine-tuning test dataset~\cite{tinyagent_dataset} as benchmarks. These queries span over various types of requests using up to a maximum of 16 different tools. Detailed configuration of experimental setup is provided in \sect{sect:methodology}. Our evaluation with the TinyAgent-7B backend LLM~\cite{tinyagent_7b} on Mac mini (M4 Pro) reveals that executing one agentic task takes 35.4 seconds on average. Even a simple task \textit{``Schedule a meeting with John tomorrow at 5pm''} takes 26.7 seconds to execute, underscoring the latency challenge for on-device agents.

\begin{figure}[t!] \centering
\includegraphics[width=\columnwidth]{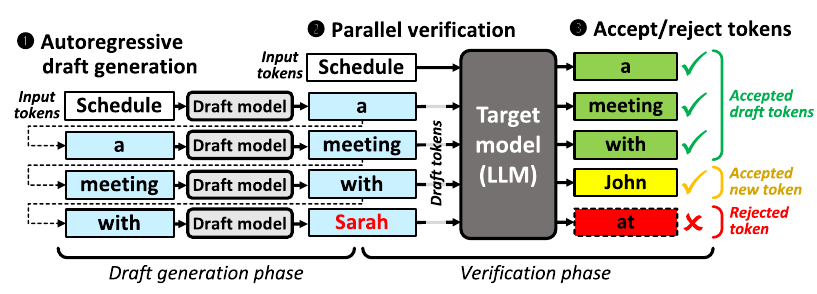}
\caption{
Illustration of speculative decoding.
}
\Description{Illustration of speculative decoding.}
\label{fig:specdec}
\end{figure}

\begin{figure}[t!] \centering
\includegraphics[width=\columnwidth]{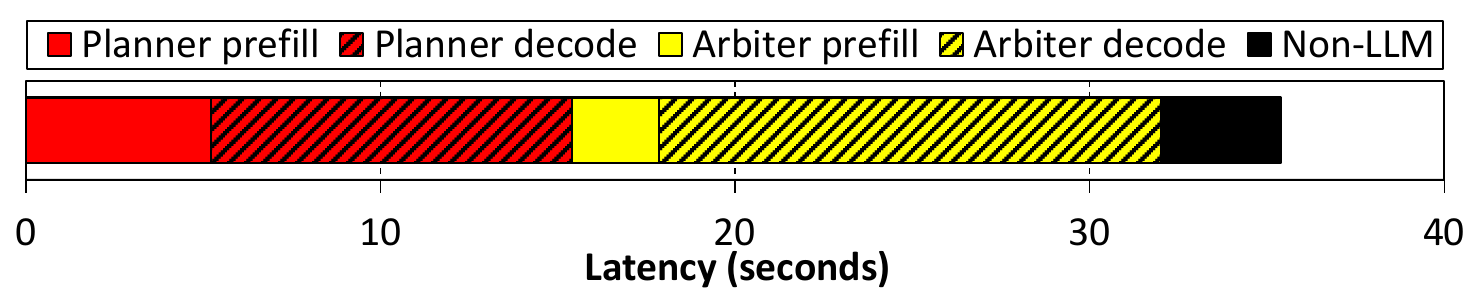}
\caption{
Latency breakdown of agentic tasks.
}
\Description{Latency breakdown of agentic tasks.}
\label{fig:agent_breakdown_latency}
\end{figure}

\fig{fig:agent_breakdown_latency} breaks down the end-to-end latency of agentic task executions. The two LLM components, Planner (43.5\%) and Arbiter (46.9\%), together account for 90.4\% of the total latency. Notably, while decode dominates (68.7\%), the prefill stage remains a significant contributor (21.7\%). This contrasts with conventional LLM workloads executed in the cloud using server-class devices, where decode latency is overwhelmingly dominant (e.g., over 95\%, 98\%, and 98\% in WSC-LLM~\cite{wsc_llm}, FDC~\cite{fdc}, and CENT~\cite{cent}), rendering the LLM decode stage the primary bottleneck to address.

We identify two main reasons for the differing bottlenecks between conventional LLMs and agentic workloads. First, prior work~\cite{gemini_2.5} reports that agents typically process much longer input token sequences than they generate as output, a trend we confirmed in our own measurements. This imbalance makes the LLM prefill stage of agents far more compute‑intensive than in conventional LLMs. Second, the cost of processing these longer inputs is magnified by limited hardware resources of today’s on‑device accelerators, rendering prefill disproportionately expensive. \tab{tab:server_vs_edge_chips} compares modern AI chips and reveals that on‑device accelerators provide at most 11\% of the memory bandwidth and roughly 2\% of the compute throughput of server‑class NVIDIA H200 GPU~\cite{nvidia_h200_sxm}. Together, these factors make the compute‑bound prefill more expensive than decode, leaving \emph{both} stages as two dominant performance bottlenecks in AI agents.

Overall, because the compute‑bound prefill and memory bandwidth‑bound decode phases now contribute comparable amounts of latency, no single stage or model instance dominates. Effective acceleration of agentic systems must therefore optimize the \textit{entire} Planner–Arbiter pipeline, improving both prefill and decode. In the remainder of this section, we present a token‑level characterization of the Planner and Arbiter prefill and decode stages that motivates our proposed \prefill and \decode.

\begin{table}[t!]
\centering
\caption{
The compute throughput and memory bandwidth available in modern AI chips, both high-end server-class devices and on-device accelerators.
}
\scriptsize
\begin{tabular}{|c|c|c|c|}
\hline
\textbf{Device} & \textbf{Class} & \makecell[c]{\textbf{Compute power}\\ \textbf{(INT8 TOPS)}} & \makecell[c]{\textbf{Memory}\\ \textbf{bandwidth (GB/s)}} \\
\hline
\textbf{NVIDIA H100}~\cite{nvidia_h100_sxm} & \multirow{5}{*}{\centering Server} & 1,979 & 3,350 \\
\cline{1-1} \cline{3-4}
\textbf{NVIDIA H200}~\cite{nvidia_h200_sxm} & & 1,979 & 4,800 \\
\cline{1-1} \cline{3-4}
\textbf{NVIDIA B200}~\cite{nvidia_b200_sxm} & & 4,500 & 8,000 \\
\cline{1-1} \cline{3-4}
\textbf{AMD MI325X}~\cite{amd_mi325x} & & 2,615 & 6,000 \\
\cline{1-1} \cline{3-4}
\textbf{Google TPU v6e}~\cite{google_tpu_v6e} & & 1,836 & 1,640 \\
\hline
\textbf{Apple M4 Max}~\cite{apple_m4_pro_max} & \multirow{5}{*}{\centering On-device} & 38 & 546 \\
\cline{1-1} \cline{3-4}
\makecell[c]{\textbf{Qualcomm Snapdragon}\\ \textbf{X Elite}~\cite{qualcomm_snapdragon_x_elite}} &  & 45 & 135 \\
\cline{1-1} \cline{3-4}
\makecell[c]{\textbf{AMD Ryzen}\\ \textbf{AI+ PRO 395}~\cite{amd_ryzen_ai_max}} &  & 50 & 256 \\
\hline
\end{tabular}
\label{tab:server_vs_edge_chips}
\end{table}

\begin{figure}[t!] \centering
\includegraphics[width=\columnwidth]{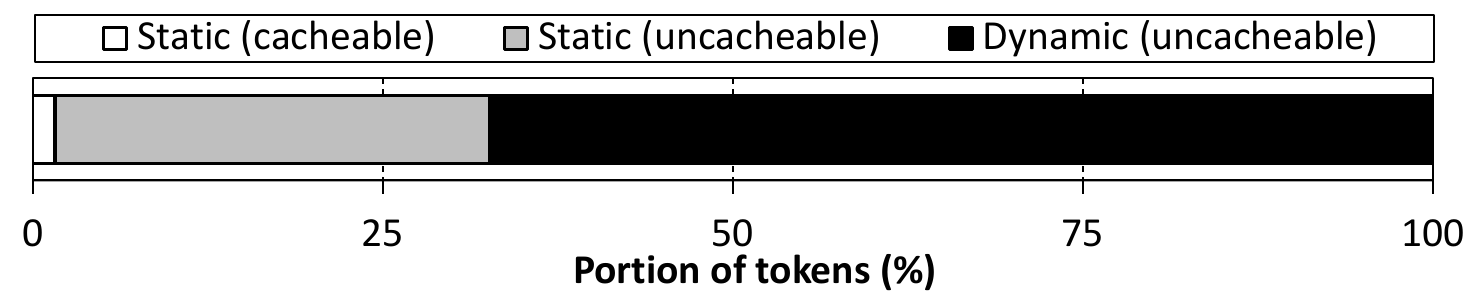}
\caption{
Token count breakdown of Planner inputs. Results are averaged across all examples in the TinyAgent fine-tuning test dataset. Static tokens are uncacheable if they are placed behind dynamic tokens.
}
\Description{Token count breakdown of Planner inputs.}
\label{fig:agent_breakdown_token}
\end{figure}

\subsection{Prefill Stage Token Analysis} \label{sect:prefill_analysis}

{\bf Planner input tokens.} The Planner input consists of the static system prompt, retrieved tool descriptions and guidelines, and retrieved tool-use examples (see \fig{fig:plan-out_agents}). \fig{fig:agent_breakdown_token} shows the average token-count distribution of Planner inputs. Out of the total 1,739 tokens, the static system prompt takes up 32.7\%, suggesting that prefix caching could potentially accelerate prefill. However, because the dynamically retrieved tool descriptions and guidelines are inserted into the Planner input early on, the first dynamic token appears after only 1.6\% of the prompt, limiting KV cache reuse if the input prompts are used as-is (see \fig{fig:prefix_caching}). Encouragingly, we observe that the dynamically changing tool descriptions and guidelines are different combinations of static fragments, where the combinations depend solely on the selected tool sets. If the tool descriptions and guidelines were made static, the number of uncacheable tokens would decrease by 32\% (1,711 to 1,171). \textit{Overall, the potential of prefix caching is hindered by early dynamicity introduced by tool descriptions and guidelines, which are merely rearranged chunks of static prompts.}

\begin{figure}[t!] \centering
\includegraphics[width=\columnwidth]{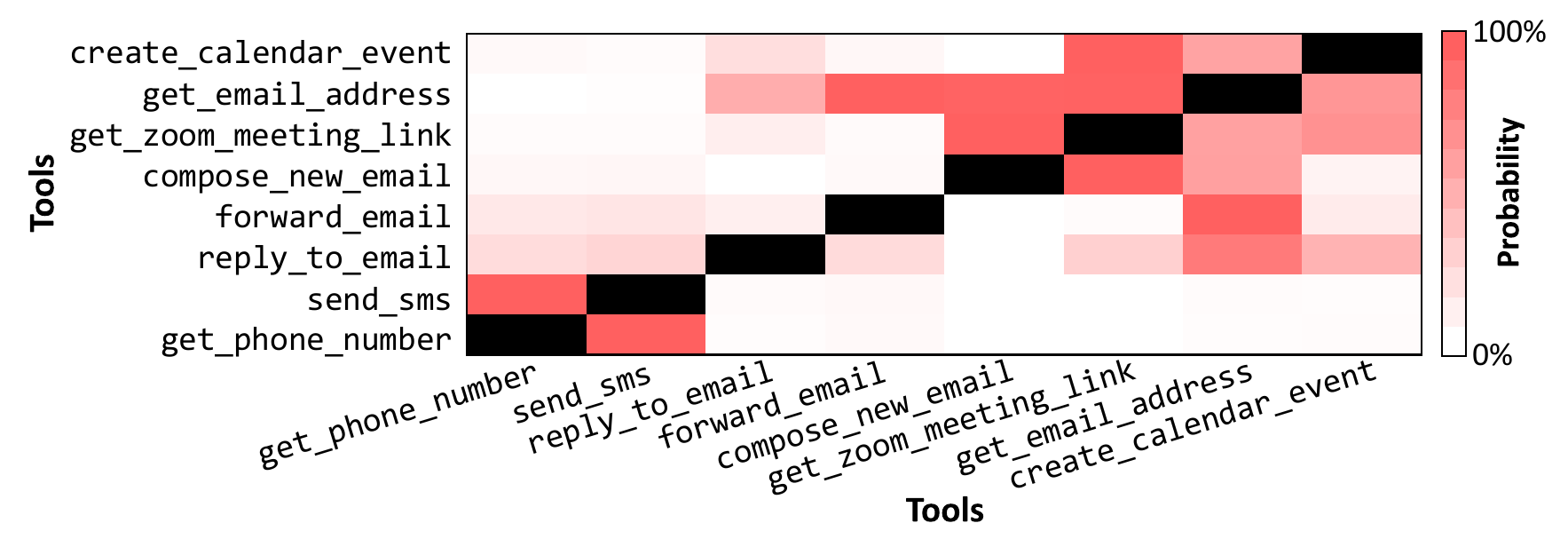}
\caption{
Tool co-activation heatmap of a subset of tools in TinyAgent training dataset. The value at $(x,y)$ depicts how likely $\text{tool}_y$ is to be activated given that $\text{tool}_x$ has been activated, i.e., $P(\text{tool}_y | \text{tool}_x)$.
}
\Description{Tool co-activation heatmap of a subset of tools in TinyAgent training dataset.}
\label{fig:tool_heatmap}
\end{figure}

We also observe the existence of \textit{tool co-activation locality} in the Planner's input prompts, which we define as the likelihood of certain tools being called together across different queries. This skewness in tool co-activation is illustrated in \fig{fig:tool_heatmap} as a heatmap. For example, \texttt{get\_zoom\_meeting\_link} is more likely to be co-activated with \texttt{get\_email\_address} (91\%) or \texttt{compose\_new\_email} (58\%) than \texttt{get\_phone\_number} (6\%). This is because the nature of agentic tasks suggests that tools which fall under the same theme (e.g., email, contacts, maps, and notes) are more likely to be retrieved together in the same plan. This locality is further extended to inter-theme relations, where certain themes like contact and email are ``closer'' in their relationship than others such as contact and maps. \textit{Thus, our characterization reveals that tool co‑activation locality is intrinsic to agentic applications.}

In the tool-use example retrieval process, the top-$K$ examples displaying the most similar text embedding to the current user query are retrieved (\sect{sect:on_device_agents}). While users typically query tasks requiring the use of multiple tools (82\% of TinyAgent training dataset), we observe that single-tool examples are retrieved as relevant examples 57\% of the time. In other words, \textit{the single-tool examples are important as tool-use examples for planning tasks.} This is not to be confused with the existence of tool co-activation locality in the choice of tools, as the dominance of single-tool examples is observed in the choice of tool-use examples.

{\bf Arbiter input tokens.} The Arbiter’s input includes decision guidelines and examples, along with the list of call-observation pairs produced by tool execution (see \fig{fig:plan-out_agents}). Depending on the internal LLMCompiler state, only two static prefix variants occur; these prefixes account for 88\% and 90\% of the Arbiter input, respectively. \textit{Because the Arbiter input contains such a large static prefix, prefix caching can capture it effectively.}

\subsection{Decode Stage Token Analysis} \label{sect:decode_analysis}

\begin{figure}[t!] \centering
\includegraphics[width=\columnwidth]{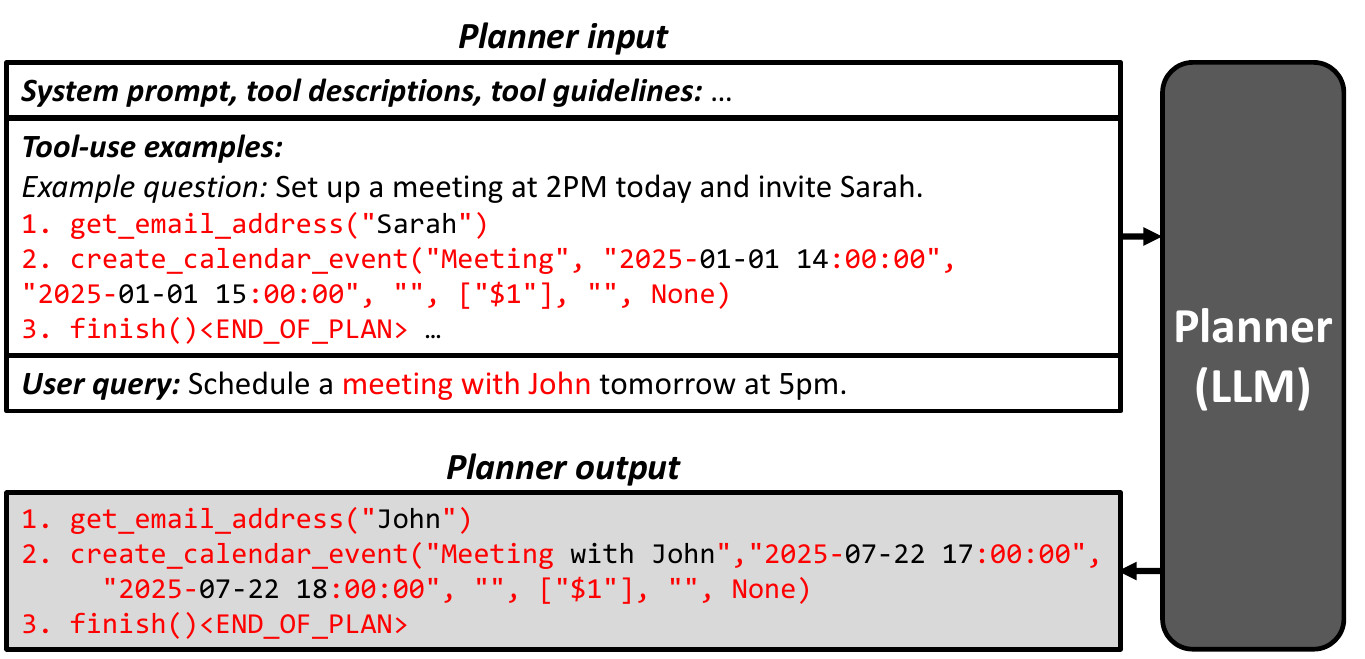}
\caption{
Input and output example of Planner for the query \textit{``Schedule a meeting with John tomorrow at 5pm.''}
}
\Description{Input and output example of Planner for the query ``Schedule a meeting with John tomorrow at 5pm.''}
\label{fig:planner_token_example}
\end{figure}

{\bf Planner and Arbiter output tokens.} \fig{fig:planner_token_example} shows an example tool-use example alongside the output for the query \textit{``Schedule a meeting with John tomorrow at 5pm.''} The generated plans share the same structural template as the tool-use example, with arguments substituted to match the user query. This occurs because the tool names, their arguments, and tool-calling orders are mostly embedded in the tool-use examples. Empirically, 96\% of Planner and 87\% of Arbiter output tokens overlap with those in their corresponding input prompt. \textit{Overall, our key observation is that the output tokens generated in both Planner and Arbiter's decode stage are highly correlated with provided examples and user prompt.}

While their outputs display highly regular and predictable patterns, reflecting the templates provided by the few-shot examples in the prompt, the decode stages of Planner and Arbiter account for 68.7\% of the end-to-end latency. Therefore, we conclude that \textit{the decode stage of agentic LLMs is inefficient in that a large portion of time is spent generating straightforward, formulaic sequences that do not require the reasoning capabilities of LLMs.}

{\bf Challenges of applying speculative decoding.} Despite its promise, speculative decoding requires a carefully chosen draft model for high efficiency, and selecting one is nontrivial. If the draft model is too small, it fails to produce high‑quality draft tokens and yields little performance improvement. Conversely, a larger draft model achieves higher token accuracy (i.e., the fraction of draft tokens ultimately accepted during verification) but introduces substantial latency overhead.

The \textit{Theoretical~max.~speedup} column of \tab{tab:draft_specdec_accuracy} reports the maximum speedup achievable by each draft model under the Planner workload. We compute this theoretical limit analytically, assuming that the LLM decode stage is memory bandwidth‑limited and that its latency therefore scales proportionally with model size. Leveraging this assumption, we combine the draft‑token accuracy (second column of \tab{tab:draft_specdec_accuracy}) to estimate the final output‑token count and its corresponding latency, which we then compare with the baseline to derive the projected speedup. As shown, even with state-of-the-art small LLMs~\cite{llama_3.2} or fine-tuned draft models from prior work~\cite{specinfer}, it is difficult to balance the accuracy and draft model size, where smaller draft LLMs barely yield any speedup due to their low draft token accuracy while large draft LLMs with high draft token latency end up spending too much time generating draft tokens.

\begin{table}[t]
\centering
\caption{
Draft token accuracy and achievable theoretical speedups of various draft models.
}
\footnotesize
\resizebox{1.0\columnwidth}{!}{
\begin{tabular}{|c|c|c|c|}
\hline
\textbf{Draft model} & \makecell[c]{\textbf{Draft token}\\ \textbf{accuracy}} & \makecell[c]{\textbf{Theoretical}\\ \textbf{max. speedup}} & \makecell[c]{\textbf{Speedup}\\ \textbf{(with tax)}} \\
\hline
\textbf{Llama-3.2-3B-Instruct} & 0.42 & 0.96$\times$ & 0.83$\times$ \\
\hline
\textbf{Llama-3.2-1B-Instruct} & 0.33 & 1.59$\times$ & 1.20$\times$ \\
\hline
\textbf{Llama-160M} & 0.02 & 0.98$\times$ & 0.57$\times$ \\
\hline
\textbf{Llama-68M} & 0.02 & 1.11$\times$ & 0.62$\times$ \\
\hline
\end{tabular}
}
\label{tab:draft_specdec_accuracy}
\end{table}

\begin{figure}[t!] \centering
\includegraphics[width=\columnwidth]{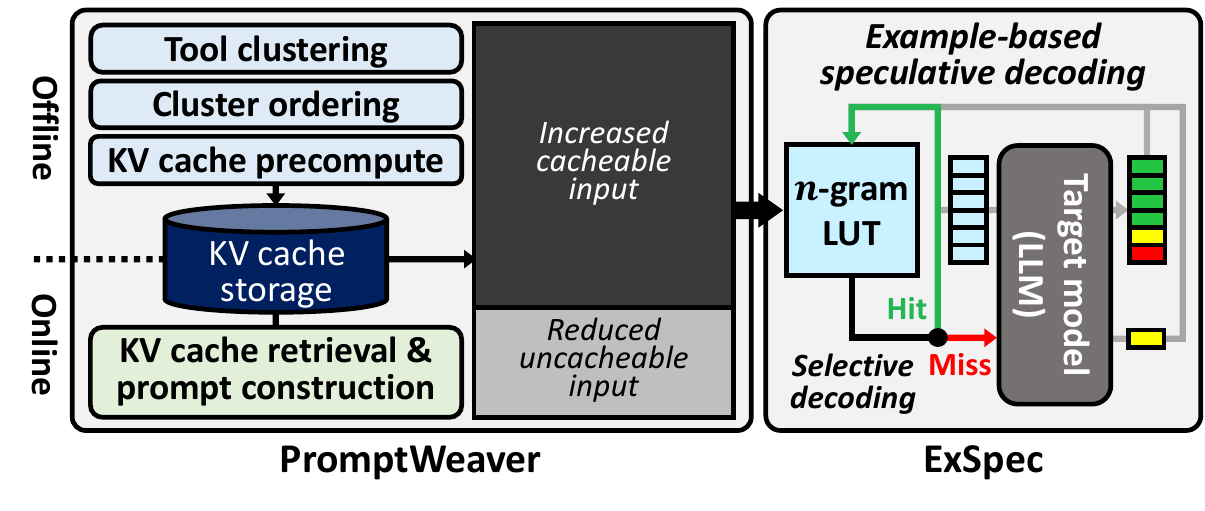}
\caption{
Proposed \proposed system architecture. 
}
\Description{Proposed \proposed system architecture. }
\label{fig:proposed_overview}
\end{figure}

Now, recall from \fig{fig:specdec} that speculative decoding requires both single‑token inference by the draft LLM (autoregressive token generation) and multi‑token inference by the target LLM (parallel verification). However, on-device AI systems are increasingly being optimized for single-batch LLM calls~\cite{decdec, llm_in_a_flash, cambricon_llm, edgemoe, edgellm}. For example, in Apple's official LLM framework MLX-LM~\cite{mlx_lm}, a single-token autoregressive inference with TinyAgent takes 131 ms per token, while the same model's verification phase with 2 tokens takes 244 ms, resulting in a 1.86$\times$ slowdown. This phenomenon, which we refer to as \textit{the multi-token tax}, degrades the overall performance of on-device speculative decoding schemes, where multiple tokens must be verified by the target LLM. Based on this slowdown, we recompute the speedup of applying speculative decoding in the \textit{Speedup~(with~tax)} column of \tab{tab:draft_specdec_accuracy}. The best achievable speedup among state-of-the-art small LLMs~\cite{llama_3.2} or draft LLMs fine-tuned by prior work~\cite{specinfer} is 1.20$\times$. \textit{Therefore, we conclude that applying speculative decoding to on-device frameworks is challenging due to the multi-token tax and draft LLM latency overhead.}
\section{\proposed: On-device AI Agents Acceleration} 
\label{sect:proposed}

\subsection{\proposed Overview} \label{sect:proposed_overview}

\fig{fig:proposed_overview} provides an overview of our \proposed system. Building on the key observations in \sect{sect:motivation}, \proposed utilizes \prefill and \decode, which respectively accelerate the prefill and decode stages of agentic systems. The objective of these techniques is to speed up their target stages without compromising task accuracy or introducing substantial overhead in the resource-constrained on-device environment.

\textit{\prefill} (\sect{sect:proposed_prefill}) reconstructs the input prompt to minimize the amount of prompt that must be computed on the fly, effectively speeding up the prefill stage. \textit{\decode} (\sect{sect:proposed_decode}) accelerates the decode stage by using a simple lookup table as the draft model for speculative decoding. This lightweight draft model incurs no draft token generation overhead, while providing means to avoid the multi-token tax. With these two main components, \proposed accelerates the full system pipeline of on-device agents, targeting both the prefill (\prefill) and the decode (\decode) stages.

\begin{figure}[t!] \centering
\includegraphics[width=\columnwidth]{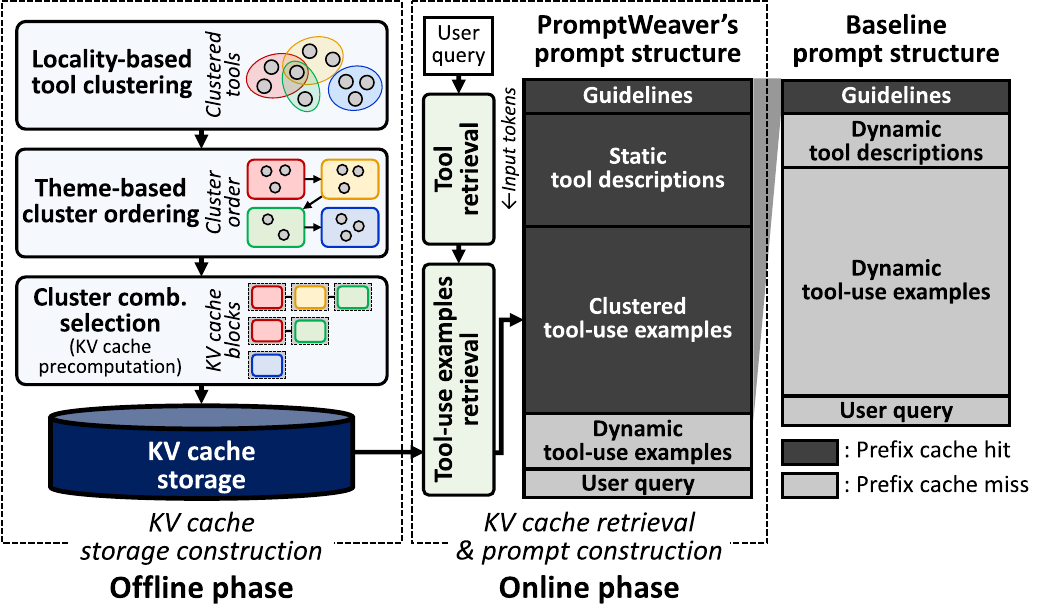}
\caption{
Overview of \prefill, divided into the offline (KV cache storage construction) and online phase (KV cache retrieval and prompt construction).
}
\Description{Overview of \prefill, divided into the offline (KV cache storage construction) and online phase (KV cache retrieval and prompt construction).}
\label{fig:proposed_prefill_overview}
\end{figure}

\begin{figure*}[t!] \centering
\includegraphics[width=1.0\textwidth]{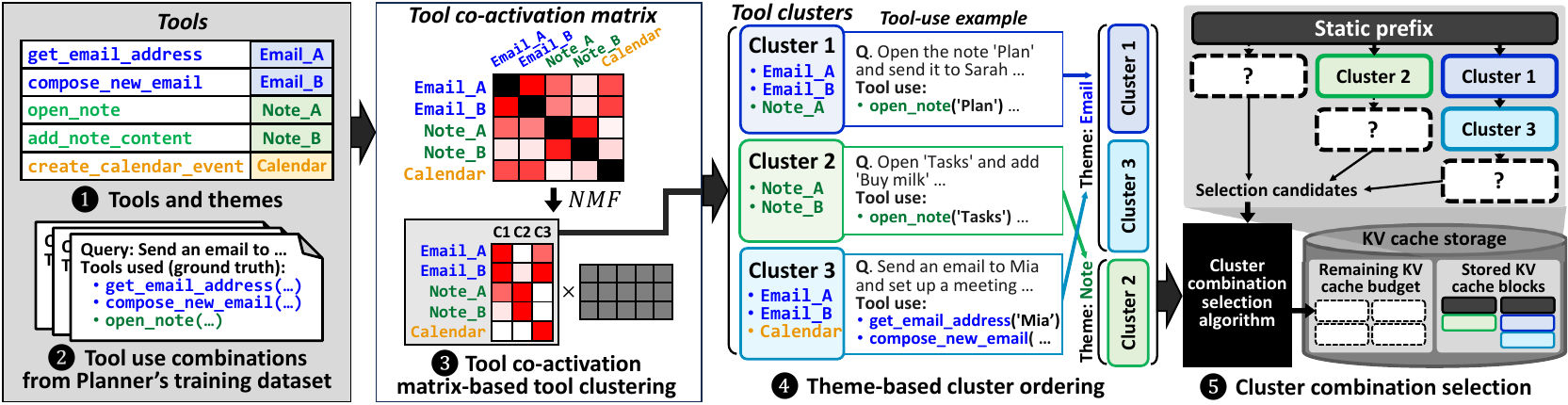}
\caption{
\label{page:fig_edit_proposed_prefill_detailed}
Illustration of \prefill's offline KV cache precompute mechanism.
}
\Description{Illustration of \prefill's offline KV cache precompute mechanism.}
\label{fig:proposed_prefill_detailed}
\end{figure*}

\subsection{\prefill: Iso-accuracy Prompt Reconstruction} \label{sect:proposed_prefill}
\label{page:iso_accuracy}

In \sect{sect:prefill_analysis}, we identified three key traits of Planner inputs: (1) early dynamicity despite a large portion of static prompt, (2) tool co-activation locality, and (3) importance of single-tool examples in tool-use examples. \prefill exploits these observations to reduce uncacheable tokens while preserving accuracy. It rebuilds prompts so that most content forms a cacheable static prefix and the dynamic regions' KV caches can be reconstructed from KV caches stored inside the SSD. \fig{fig:proposed_prefill_overview} provides an overview of how \prefill introduces staticity without degrading accuracy. The reconstructed prompts consist of the following elements, in order: (1) all-inclusive static tool descriptions and guidelines, (2) clustered semi-cacheable tool-use examples, and (3) uncacheable dynamic tool-use examples.

{\bf Replacing early dynamic tokens.} The first dynamic fragment of prompt contains descriptions and usage guidelines for the selected tools. Because the set of tools selected depends on the input prompt and is thus determined at runtime, this section is where dynamicity is introduced. \prefill reconstructs this section to include descriptions and guidelines for \textit{all tools} available in the system. This replacement transforms the early dynamic segment into a larger but fully static prefix. Although this increases the size of the KV cache, it enables the prompts made static to be \emph{precomputed} and have their KV caches stored in the SSD. Since static prefixes can be cached once and reused across requests, our approach significantly reduces the number of uncacheable tokens in the prompt. Concretely, tokens previously marked as ``Static (uncacheable)'' in \fig{fig:agent_breakdown_token} are now part of the cacheable prefix, thereby reducing the dynamic token volume at runtime at the cost of a larger KV cache footprint.

In the process of inducing staticity, the input prompt grows longer. The average length of a single tool's description and guidelines is 120 tokens. For an agentic system using up to $t$ tools, including all descriptions and guidelines requires $120 \cdot t$ tokens. Even with a conservative estimate of 100 tools, our all-inclusive static tool description adds only 1.4 GB of additional KV cache stored on the SSD. In general, the performance benefits of our approach far outweigh the overhead introduced by this larger KV cache.

{\bf Tool-use example selection.} The remaining uncacheable input tokens primarily consist of tool-use examples. These examples are dynamic because they are selected at runtime based on the user prompt (\sect{sect:on_device_agents})~\cite{toolformer, rag_mcp, hugginggpt}. The all-inclusive static tool description and guidelines cannot be applied to tool-use examples for two reasons. First, supplying the Planner with all possible combinations of tool-use examples requires $2^t-1$ different combinations for a system with $t$ tools. For the current TinyAgent with 16 tools, this is equivalent to 800 GB of memory, increasing exponentially as more tools are supported ($memory \propto 2^t$). Second, the few-shot examples (i.e., tool-use examples) serve as templates and have a direct impact on the generation quality of LLMs. Adding such an excessive amount of tool-use examples would hamper the Planner's ability to extract relations between different tools~\cite{toolformer, rag_mcp, google_dev_function_calling}. Instead, \prefill chooses a fixed set of tool-use examples that can be reused across different queries at the offline phase through (1) co-activation locality-based tool clustering, (2) theme-based cluster ordering, and (3) cluster combination selection, as illustrated in \fig{fig:proposed_prefill_detailed}.

{\bf Co-activation locality-based tool clustering.}
In this stage, tools are clustered based on their co-activation locality (\fig{fig:tool_heatmap}). First, the tools are annotated with themes that relate to the tool's usage (\circlednum{1}), information that \prefill utilizes in the next theme-based cluster ordering stage. Then, \prefill refers to the Planner's training dataset (TinyAgent fine-tuning training dataset, \sect{sect:workload_characterization}) and iterates through all training data samples' ground truth labels to identify which tools are called together (\circlednum{2}). With this information, we generate a tool co-activation matrix, which is an adjacency matrix where each graph node corresponds to a tool and edges denote the number of times the corresponding pair of tools has been called together, and apply non-negative matrix factorization (NMF)~\cite{nmf} to cluster the tools based on the co-activation matrix (\circlednum{3}). This results in a total of eight clusters, each consisting of 2 to 6 tools. After the tools are clustered, we assign each cluster one tool-use example that uses exactly the tools in that cluster. These clusters then undergo theme‑based ordering and cluster combination selection (detailed later) before being stored in SSD for retrieval during inference. For example, when servicing a user query, ToolRAG retrieves the set of tools likely to be used for the request (\fig{fig:toolrag}(b)). A cluster is considered ``activated'' if it contains at least one of these tools, and the corresponding tool‑use examples from the activated clusters are included as few‑shot examples in the prompt (``Clustered tool‑use examples'' in \fig{fig:proposed_prefill_overview}).

{\bf Theme-based cluster ordering.}
To maximize KV cache reuse, clusters must be given a fixed ordering so that when the same combination of clusters are activated, the resulting KV cache yields the same value. That is, because having clusters A and B in order ``A-B'' and ``B-A'' yield different KV cache values, we impose a fixed ordering so that between clusters A and B, the order is always statically fixed as ``A-B'' (or ``B-A''). To maximize the amount of KV cache reuse under a fixed SSD capacity budget, \prefill applies theme-based clustering and cluster combination selection. The objective of theme‑based ordering is to place clusters that are frequently co‑activated adjacent to each other. Each cluster is assigned a theme based on which tool theme is most dominant in the cluster. Then, clusters with identical themes are grouped together and placed next to each other in a fixed order (\circlednum{4}). For example, clusters with the same theme ``Email'' are grouped together and placed adjacent to each other in order. If the tool \texttt{get\_email\_address} (\texttt{Email\_A}) is retrieved, clusters \texttt{1} and \texttt{3}, which include the tool, are activated. When the clusters are ordered based on their themes (after ordering in \circlednum{4}), having stored the KV cache of clusters ``\texttt{1}-\texttt{3}-\texttt{2}'' would fully cover for tool \texttt{get\_email\_address}, because if cluster \texttt{2} is not activated, we can cut out the mismatched tail (cluster \texttt{2}) and reuse the KV cache of ``\texttt{1}-\texttt{3}.'' However, if clusters were not grouped by themes and put in random order (before ordering in \circlednum{4}), it would require two cluster combinations (``\texttt{1}-\texttt{2}-\texttt{3}'' and ``\texttt{1}-\texttt{3}'') depending on whether or not cluster \texttt{2} is activated. Thus, with a fixed, theme-based ordering, \prefill can maximize prefix cache reuse across various requests.

\begingroup
\begin{algorithm}[t]
\caption{Cluster combination selection}
\label{algo:greedy_selection}
\begin{algorithmic}[1]
\footnotesize
  \STATE \textbf{Input:} KV cache budget $N$, Planner's training dataset $\mathcal{D}$
  \STATE \textbf{Output:} cluster combination KV cache $\mathcal{C}$
  \STATE Initialize $\mathcal{C} \gets \emptyset$
  \STATE Initialize prefixes $P \gets \{\text{all prefixes of sequences in }\mathcal{D}\}$
  \FOR{$i = 1$ to $N$}
    \STATE $\mathrm{options}\gets \emptyset$
    \FORALL{prefix $p$ in $P$}
      \IF{len($p$) == 1 or $p$[:-1] in $\mathcal{C}$}
        \STATE options.add($p$)
      \ENDIF
    \ENDFOR
    \STATE $\hat{p}=\displaystyle\argmax_{p \in \mathrm{options}} \left[ \mathrm{coverage}(\mathcal{D}, \mathcal{C}\cup\{p\}) {-} \mathrm{coverage}(\mathcal{D}, \mathcal{C}) \right]$
    \STATE $\mathcal{C}$.add($\hat{p}$)
  \ENDFOR
  \STATE \textbf{return} $\mathcal{C}$
\end{algorithmic}
\end{algorithm}
\endgroup

{\bf Cluster combination selection.}
The total number of possible cluster combinations is $2^C-1$ for $C$ ordered clusters. \prefill must carefully select combinations of clusters to store in the SSD to maximize KV cache reuse across requests, while keeping the SSD capacity overhead low. \algo{algo:greedy_selection} explains our proposed combination selection algorithm. Given a cache budget of $N$ clusters, starting from an empty KV cache set $\mathcal{C}$, all available cluster prefixes of all data samples in the Planner's training dataset ($\mathcal{D}$, same one used in locality-based clustering) are gathered (line 4). That is, if a sample activates clusters ``A-B-C'', its prefix sequences ``A,'' ``A-B,'' and ``A-B-C'' are all gathered. We define a new metric to give scores for each combination of cached clusters $\mathcal{C}$. For each activated cluster sequence in $\mathcal{D}$, the length of the longest cached prefix (i.e., how many leading clusters can reuse the KV caches in $\mathcal{C}$) is measured. The \textit{coverage} is defined by the sum of these per-sequence number of hit clusters in $\mathcal{D}$. A larger coverage implies a higher KV cache reuse across requests. Each step considers all candidate prefixes, either a singleton cluster (len($p$) == 1 in line 8, first dotted box in \circlednum{5}) or an extension to an existing sequence by one cluster of a prefix already owned ($p$[:-1] in $\mathcal{C}$ in line 8, second and third dotted boxes in \circlednum{5}), and computes how much coverage would increase if that candidate were added to the cached set (line 12). By greedily choosing the prefix that adds the largest amount of new coverage, the algorithm focuses on the most frequent early patterns without exploring every combination. The returned cluster combinations $\mathcal{C}$ have their KV cache precomputed and saved in the SSD for reuse at the online phase (\circlednum{5}). In \sect{sect:eval_prefill}, we show that with a cache budget size of just 15 clusters (5.87 GB of SSD capacity overhead), 74.4\% of tool-use examples are covered. This demonstrates that a small, carefully chosen subset of cluster combinations can serve the vast majority of prompt patterns, dramatically reducing SSD capacity overhead while preserving high KV cache reuse.

{\bf Preserving task accuracy with dynamic tool-use examples.} LLM output quality is known to be heavily dependent on the quality of few-shot examples in the prompt~\cite{gpt3, what_makes_good_examples, metalcl}. Therefore, composing the entirety of tool-use examples with clustered examples can have critical impact on the Planner accuracy. To alleviate this, we append single-tool examples\footnote{We add double-tool examples for a select few tools whose single-tool example was not present in the offline-generated database.} of activated tools, which we observed to be the most popular form of tool-use examples, to the end of the clustered examples. We also add top-$K$ ($0 \le K \le 4$) relevant examples from ToolRAG (\fig{fig:toolrag}(c)) to make up for any accuracy loss. In \sect{sect:eval_prefill}, our evaluations reveal that $K=1$ is the optimal choice in terms of accuracy, requiring only one additional tool-use example worth of dynamic tokens.

\subsection{\decode: Example-based Selective Speculative Decoding} \label{sect:proposed_decode}

\begin{figure}[t!] \centering
\includegraphics[width=\columnwidth]{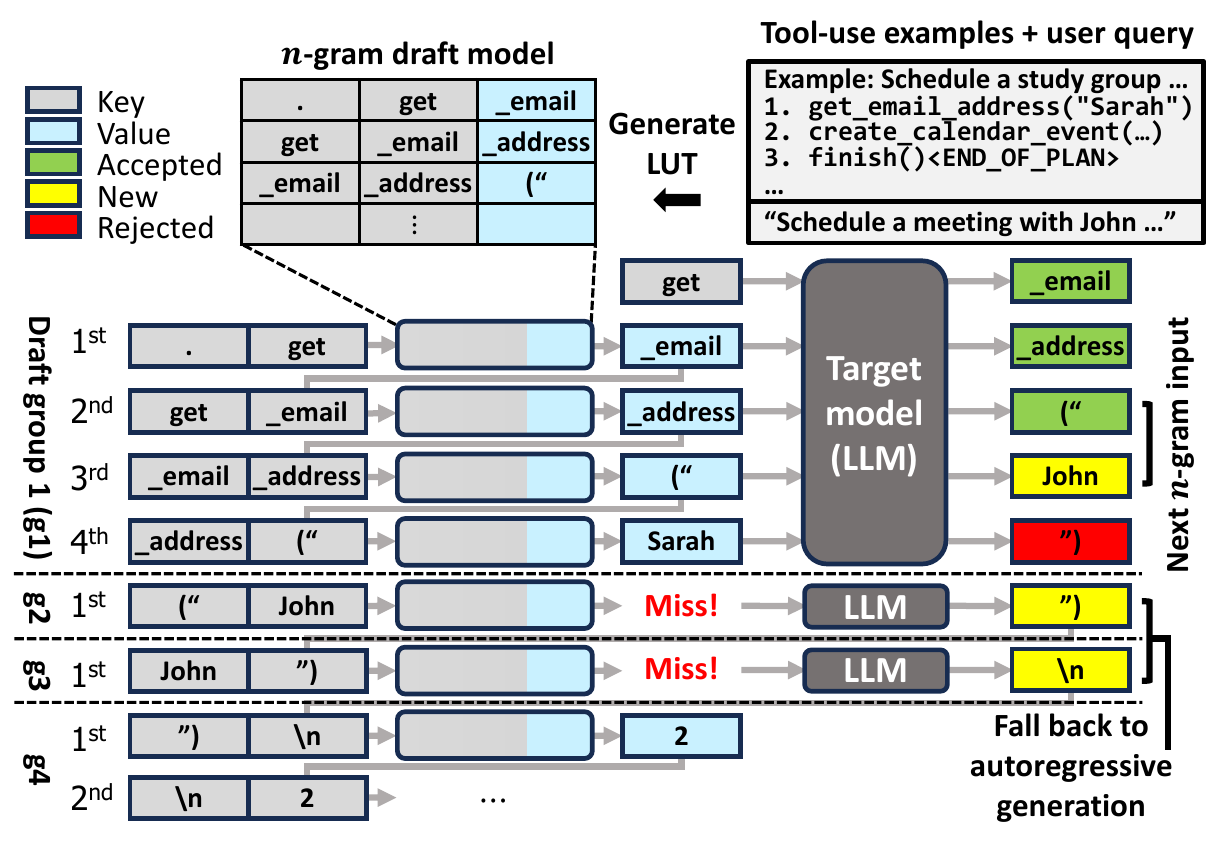}
\caption{
Overview of \decode with trigram ($n{=}3$) LUT and draft token generation length of 4.
}
\Description{Overview of \decode with trigram ($n{=}3$) LUT and draft token generation length of 4.}
\label{fig:proposed_decode_overview}
\end{figure}

\sect{sect:decode_analysis} uncovered two key properties of Planner and Arbiter output tokens: (1) decoded outputs are largely predictable from input prompts, and (2) applying speculative decoding na\"ively in on-device frameworks leads to suboptimal performance due to penalties from the multi-token tax and draft LLM latency overheads. We propose \decode to leverage this predictability without incurring these penalties. \fig{fig:proposed_decode_overview} illustrates the two main mechanisms of \decode: (1) lightweight draft token generation using an $n$-gram model~\cite{ngram} and (2) selective fallback to autoregressive generation when speculative decoding is likely to be inefficient.

{\bf Draft LLM-free, example-based speculative decoding.}
A primary design objective of \decode is to remain lightweight while keeping decoded outputs tightly correlated with the input prompt. To this end, \decode builds a simple $n$‑gram lookup table (LUT) on the fly each time the agent receives a new user query. The LUT is populated as follows. To keep this table task-specific, we build the table from the few-shot examples (tool-use examples for Planner and decision examples for Arbiter) and the user query, all of which constitute a single stream of tokens. We then slide a window of $n$ consecutive tokens $t_{1:n}$ across this token stream, shifting one token at a time, and record each pair $\langle t_{1:n-1}, t_{n} \rangle$. The prefix $t_{1:n-1}$ serves as the key in the $n$‑gram LUT, and the corresponding value is the token $t_{n}$ that occurs most frequently with that key across all recorded pairs. Using this LUT, \decode generates each draft token by indexing the table with the most recent $n-1$ generated tokens (gray tokens in \fig{fig:proposed_decode_overview}) and the corresponding output becomes the draft token (blue token). A key advantage of \decode's LUT design is its minimal memory footprint, only amounting to a few KB. This lightweight design is ideal for on‑device AI systems, unlike conventional LLM‑based drafts, which consume hundreds of MB to several GB of memory (\tab{tab:draft_specdec_accuracy}).

Balancing $n$ is crucial for maximizing \decode's performance. With a small $n$, for example a unigram model ($n=1$), the LUT always proposes the single most frequent token from the few‑shot examples and user query, yielding low‑quality draft tokens. Conversely, a large $n$ provides richer context and higher‑quality drafts but often fails to predict unseen sequences, defaulting to random tokens when the $n-1$ prefix is not available in the LUT. We find that a trigram model ($n=3$) offers the best trade‑off between draft quality and LUT hit rate. The impact of $n$ is further explored in \sect{sect:eval_discussions}.

\begin{figure}[t!] \centering
\includegraphics[width=\columnwidth]{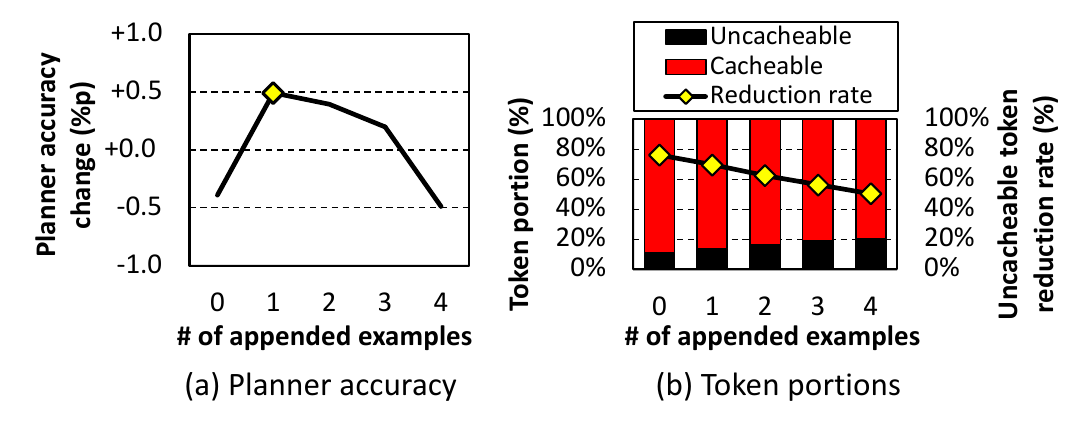}
\caption{
\label{page:fig_clarification_eval_planner}
(a) Planner accuracy change vs. the number of appended tool-use examples.
(b) Share of cacheable and uncacheable tokens (left) and reduction in uncacheable tokens (right).
}
\Description{(a) Planner accuracy change vs. the number of appended tool-use examples.
(b) Share of cacheable and uncacheable tokens and reduction in uncacheable tokens.}
\label{fig:planner_accuracy_tokens}
\end{figure}

{\bf Multi‑token‑tax evasion with selective decoding.}
When the current $(n-1)$-token context is absent from the LUT, our $n$-gram draft model ``guesses'' a random token that rarely passes verification. Verification still generates the same output tokens via standard autoregressive decoding and incurs the multi‑token tax, ultimately slowing down overall token generation. \decode mitigates this limitation by opting out of speculative decoding when no speedup is expected. Specifically, \decode consults the LUT before generating the first draft token: if no valid entry exists, it immediately falls back to standard autoregressive decoding (yellow tokens in \fig{fig:proposed_decode_overview}). This safeguard ensures speculative decoding is used only when its speedup outweighs the verification cost. Because the $n$‑gram LUT deterministically knows which contexts it covers, this decision incurs zero overhead—an assurance LLM‑based drafts cannot provide. Once the first draft token is retrieved from the LUT, the remaining tokens in that group are generated regardless of subsequent misses.
\section{Evaluation} \label{sect:eval}

\subsection{Methodology} \label{sect:methodology}
{\bf Model and dataset.} We target TinyAgent~\cite{tinyagent}, an open-source, on-device agent framework for macOS built on LLMCompiler~\cite{llmcompiler}. Our experiments use TinyAgent-7B~\cite{tinyagent_7b} as the backend LLM, a fine-tuned variant of WizardLM-2-7B~\cite{wizardlm2}. TinyAgent’s fine-tuning dataset~\cite{tinyagent_dataset} serves three roles: \prefill uses the training split for tool clustering and combination selection, and the test split is used for evaluation. Before each task, we flush the system’s page cache to isolate the cost of loading the KV cache from storage.

{\bf Hardware and software.} Experiments run on an Apple Mac mini with an M4 Pro chip~\cite{apple_m4_pro_max}, 64 GB of memory, 512 GB of SSD storage, 12 CPU cores, and 16 GPU cores. We build \prefill and \decode on MLX-LM~\cite{mlx_lm} and MLX-engine~\cite{mlx_engine}. MLX-LM is Apple Silicon’s official LLM inference package, backed by MLX~\cite{mlx}. MLX-engine is an open-source MLX LLM engine from LM Studio. The software stack comprises MLX v0.25.2, a modified MLX-LM v0.25.1, and MLX-engine commit \#ecc2cf4 on macOS Sequoia 15.5.

\subsection{\prefill} \label{sect:eval_prefill}

{\bf Planner accuracy.}
In line with prior work~\cite{llmcompiler,tinyagent}, we define the Planner accuracy by constructing a Directed Acyclic Graph (DAG) from the output plan, where each node represents a function call and a directed edge represents the dependency, and compare the generated DAG against the ground truth's DAG. Because the tool calls are deterministic, the Planner accuracy directly translates to the end-to-end task accuracy of the agent. \prefill adds $K (0 \le K \le 4)$ tool-use examples to make up for its accuracy loss. \fig{fig:planner_accuracy_tokens}(a) compares the Planner's task accuracy changes with the baseline as more examples are added. Not surprisingly, $K=0$ results in lower accuracy (0.832), lower than baseline's 0.836. We observe that the accuracy peaks at $K=1$ (0.841) and falls as more examples are added.

\fig{fig:planner_accuracy_tokens}(b) shows how the share of cacheable vs. uncacheable tokens changes as tool-use examples are added. Each new example is uncacheable, so the uncacheable fraction grows from 11\% at $K{=}0$ to 21\% at $K{=}4$. Accuracy peaks at $K{=}1$, so we adopt this setting for \prefill. With $K{=}1$, \prefill averages 519 uncacheable tokens, a 70\% drop from the baseline's 1,711. This result highlights \prefill's ability to minimize uncacheable tokens.

\begin{figure}[t!] \centering
\includegraphics[width=\columnwidth]{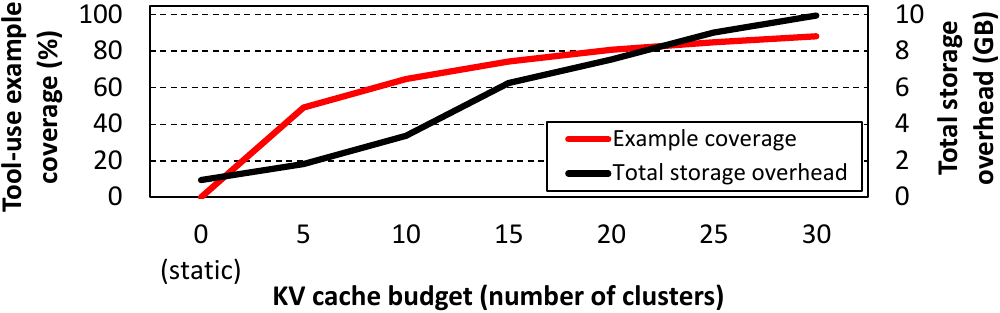}
\caption{
Tool-use example coverage (left, red) and total storage overhead (right, black) by KV cache budget.
}
\Description{Tool-use example coverage and total storage overhead by KV cache budget.}
\label{fig:eval_clustering}
\end{figure}

{\bf Storage overhead.}
Precomputed KV caches are stored in SSD, and loaded to memory on demand to reduce online prefill compute. Because the agentic workloads display long prompt lengths, the size of the KV cache is also large. Expanding the cluster budget boosts tool-use example coverage (\sect{sect:proposed_prefill}) but also raises storage cost, as shown in \fig{fig:eval_clustering}. At budget 0, only static tokens are cached, using 0.95 GB (0.57 GB, Planner + 0.39 GB, Arbiter). Coverage grows with larger budgets but levels off beyond 15 clusters. We therefore fix the budget at 15 clusters for the remainder of the evaluation, using 6.26 GB of storage for 74.4\% coverage.

\begin{figure}[t!] \centering
\includegraphics[width=\columnwidth]{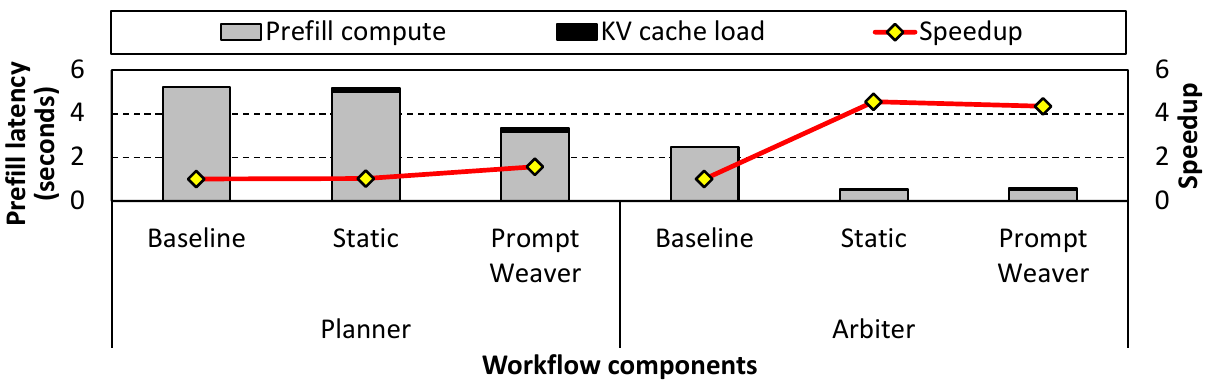}
\caption{
Prefill stage latency (left) and speedup gain (right) of \prefill.
}
\Description{Prefill stage latency and speedup gain of \prefill.}
\label{fig:eval_prefill_speedup}
\end{figure}

{\bf Speedup analysis.} 
\fig{fig:eval_prefill_speedup} quantifies impact of \prefill on Planner and Arbiter prefill stages. The Static design, caching only static tokens for the prompt, adds just a 1.01$\times$ speedup for Planner because while the amount of static tokens increases, the number of uncached dynamic tokens stays roughly the same (15.3\% decrease). In \prefill, clustered dynamic caching is applied on top of static token caching, achieving a 49.6\% reduction in the number of uncached input tokens and a 1.57$\times$ speedup. Meanwhile, \prefill reduces the amount of uncached input tokens by 88.9\% thanks to Arbiter's mostly static input tokens, achieving 4.35$\times$ speedup. Overall, loading KV cache from SSD storage adds minor overhead compared to the speedup achieved by compute savings, accounting for 5.8\% and 11.7\% of prefill latency in Planner and Arbiter, respectively.

\subsection{\decode} \label{sect:eval_decode}

\begin{figure}[t!] \centering
\includegraphics[width=\columnwidth]{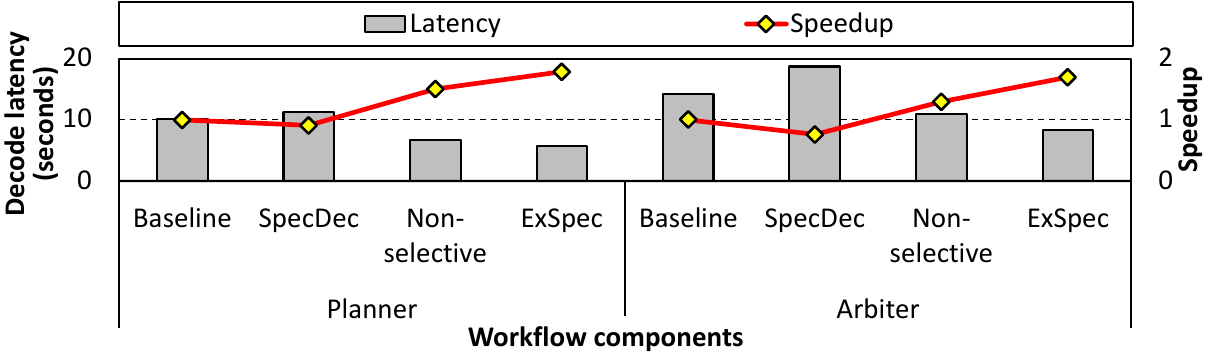}
\caption{
Decode stage latency (left) and speedup (right) of speculative decoding (SpecDec) and \decode.
}
\Description{Decode stage latency and speedup of speculative decoding (SpecDec) and \decode.}
\label{fig:eval_decode_speedup}
\end{figure}

{\bf Speedup analysis.}
\fig{fig:eval_decode_speedup} reports the decode stage latency of speculative decoding with draft LLM Llama-3.2-1B-Instruct~\cite{llama_3.2} (SpecDec) and \decode with a trigram ($n{=}3$) draft model. SpecDec experiences slowdown rather than speedup over the baseline. Other than the multi-token tax, the overhead that comes from dealing with different tokenizers between the target and the draft model~\cite{hf_uag} further slows down the system to achieve a much slower speed than was expected from \tab{tab:draft_specdec_accuracy}. On the contrary, the non-selective design of \decode effectively avoids any draft LLM-related overheads, speeding up the decode stage by 1.38$\times$. With selective decoding, \decode can avoid the multi-token tax when no draft tokens are accepted, falling back to autoregressive generation 17 (Planner) and 37 (Arbiter) times per query. Overall, \decode reduces the decode latency by 1.73$\times$, establishing itself as a lightweight and performant solution.

{\bf Draft token accuracy.}
\tab{tab:eval_draft_token_accuracy} compares draft-token accuracy under selective and non-selective \decode. Although non-selective decoding produces many more draft tokens, both modes accept the same number, underscoring the value of reverting to autoregressive generation when acceptance chances are low. Selective decoding thus increases overall accuracy, falling back to autoregressive generation on average 17 (Planner) and 37 (Arbiter) times per query.

{\bf $\bm{n}$-gram model generation overhead.}
While the $n$-gram model incurs constant-time latency per lookup, building the LUT takes $O(N)$ time for input token length $N$. The LUT generation takes up 83 milliseconds per query, indicating that it incurs a negligible overhead in terms of latency.

\subsection{\proposed Full System Integration} \label{sect:eval_end_to_end}

\begin{figure}[t!] \centering
\includegraphics[width=\columnwidth]{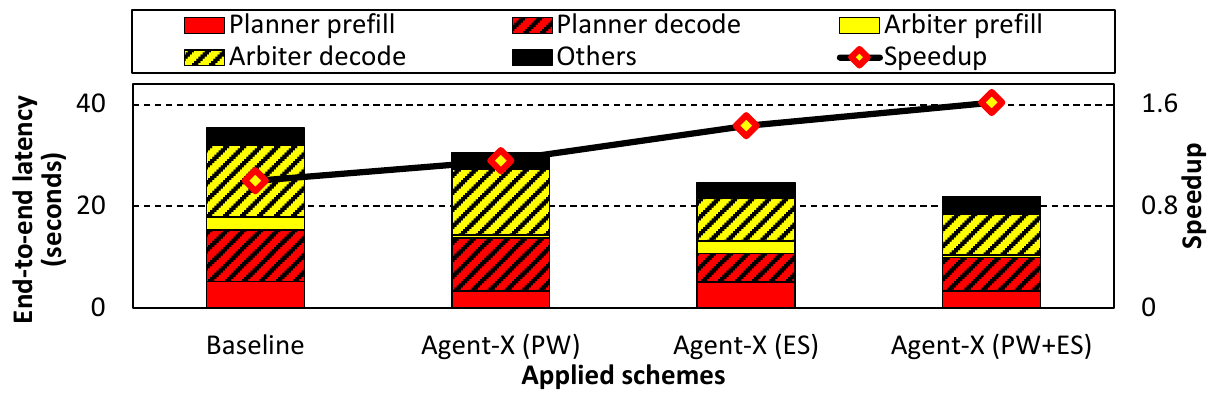}
\caption{
End-to-end latency (left) and speedup (right) of \proposed. PW denotes \prefill, ES denotes \decode, and PW+ES denotes the full \proposed pipeline.
}
\Description{End-to-end latency and speedup of \proposed.}
\label{fig:eval_end_to_end_speedup}
\end{figure}

{\bf End-to-end speedup.}
\fig{fig:eval_end_to_end_speedup} shows the latency and speedup of end-to-end on-device agentic workloads. The impacts of \prefill (PW), \decode (ES), and \proposed (PW+ES) are shown. Applying \prefill and \decode independently provides end-to-end speedups of 1.16$\times$ and 1.43$\times$, respectively. Applying \prefill and \decode together reaps an end-to-end speedup of 1.61$\times$. Alongside the speedup, it is worth noting that \proposed is a purely software solution that can be directly applied to existing on-device agentic systems without degrading the accuracy of the agent.

\begin{table}[t]
\centering
\caption{
Draft token accuracy comparison between non-selective and selective decoding schemes of \decode.
}
\footnotesize
\resizebox{1.0\columnwidth}{!}{
\begin{tabular}{|c|c|c|c|c|}
\hline
\makecell[c]{\textbf{Workflow}\\ \textbf{component}} & \makecell[c]{\textbf{Applied}\\ \textbf{method}} & \makecell[c]{\textbf{Generated}\\ \textbf{draft tokens}} & \makecell[c]{\textbf{Accepted}\\ \textbf{draft tokens}} & \makecell[c]{\textbf{Draft token}\\ \textbf{accuracy}} \\
\hline
\multirow{2}{*}{\centering Planner} & Non-selective & 364 & 48 & 0.13 \\
\cline{2-5}
 & Selective & 194 & 48 & 0.25 \\
\hline
\multirow{2}{*}{\centering Arbiter} & Non-selective & 622 & 56 & 0.09 \\
\cline{2-5}
 & Selective & 218 &56 & 0.26 \\
\hline
\end{tabular}
}
\label{tab:eval_draft_token_accuracy}
\end{table}

\subsection{Discussions} \label{sect:eval_discussions}

{\bf Application to other platforms, agents, and models.}
As \prefill and \decode are both purely software solutions, \proposed can easily be ported to other hardware platforms. While our prototype is implemented on macOS due to the maturity of its local LLM ecosystem (TinyAgent), the core algorithms of \proposed are hardware-agnostic and applicable to other platforms. Because the fine-tuned model and datasets provided by TinyAgent are available only in macOS environments, an end-to-end evaluation of \proposed on other hardware platforms is challenging. We also evaluate \proposed on the smaller model TinyAgent-1.1B. \prefill and \decode achieve speedups of $1.62\times$ (prefill) and $1.42\times$ (decode), confirming their efficacy on smaller LLMs.

{\bf Fine-tuned models as draft LLM.}
We explore using TinyAgent-1.1B, fine-tuned for agentic tasks, as the draft model in applying speculative decoding to target TinyAgent-7B. Our experiments under the same setting as \fig{fig:eval_decode_speedup} show it is 1.81$\times$ slower than the baseline. Its draft token accuracy displays high variance across tasks, exhibiting unstable latency. Based on the TinyAgent paper~\cite{tinyagent}, we estimate its fine-tuning cost as 5 ExaFLOPs ($5\times10^{18}$ FLOPs), and even at 100\% compute utilization, it would take 75 hours to fine-tune TinyAgent-1.1B on our evaluation platform. Overall, the added fine-tuning cost and the resulting latency slowdown demonstrate that using a fine-tuned draft LLM is not beneficial, underscoring the effectiveness of \decode.

{\bf Robustness to tool-use distribution drift.}
To assess the effect of tool-use pattern drifts on \prefill, we evaluate a case where notes-related tools, namely \texttt{create\_note}, \texttt{open\_note}, and \texttt{append\_note\_content}, are disabled. \prefill only experiences a small accuracy drop (0.8\%p) and maintains high tool-use example coverage with 75.7\% at cache budget of 15 clusters. Overall, these numbers are close to those reported in \fig{fig:planner_accuracy_tokens} and \fig{fig:eval_clustering}, demonstrating the robustness of \prefill.

{\bf Effect of lengthened input prompt on decode latency.}
\prefill increases the average input tokens from 1,739 to 3,790 to enable prefix caching. This translates to an additional 256 MB of memory for the KV cache, forcing the memory bandwidth-bound decode stage to load more data. Consequently, the normalized decode stage latency, quantified as \emph{Time-Per-Output-Token}, increases by 2.2\% (from 122 ms to 125 ms), consistent with a 1.7\% rise in overall memory usage. However, the prefill speedup outweighs this minor decode overhead. Furthermore, additional tokens only apply to the Planner, reducing its impact on the full system pipeline.

\begin{figure}[t!] \centering
\includegraphics[width=\columnwidth]{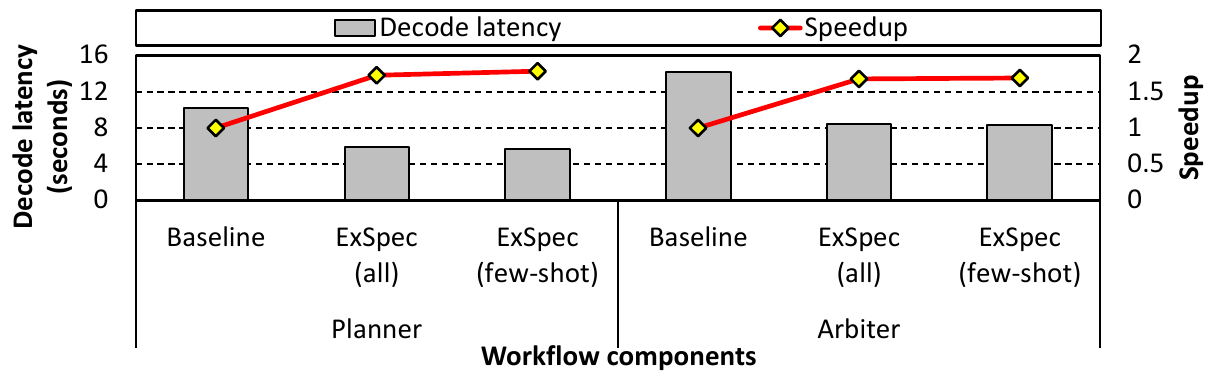}
\caption{
Decode latency (left) and speedup (right) of \decode with varying extraction regions for LUT.
}
\Description{Decode latency and speedup of \decode with varying extraction regions for LUT.}
\label{fig:discussion_extraction_region}
\end{figure}

{\bf Prompt extraction region.}
\decode constructs the $n$-gram LUT using a portion of the input. To study the effect of the extraction region, we compare two cases: \decode~(all), which uses the entire input, and \decode~(few-shot), our proposed method, which extracts the few-shot examples and the user query. As shown in \fig{fig:discussion_extraction_region}, \decode~(all) still achieves a solid 1.70$\times$ speedup over the baseline. However, \decode~(few-shot) delivers an additional 3\% and 1\% speedup over \decode~(all) for the Planner and Arbiter, respectively. This indicates that the Planner is more sensitive to the choice of extraction region than the Arbiter. We attribute this sensitivity to the input size, where the Planner is 2.20$\times$ larger than Arbiter, therefore ``polluting'' the LUT with excess tokens.

{\bf Various \texorpdfstring{$n$}{n}-gram draft models.}
We examine how $n$ in \decode's $n$-gram model affects its effectiveness. A bigram model ($n{=}2$) causes the draft token accuracy to drop sharply to 0.10, a significant decrease compared to the default ($n{=}3$), which achieves 0.25. This result highlights the importance of the context length ($n{-}1$) in our $n$-gram draft model in generating accurate predictions. A quadgram model ($n{=}4$) further improves accuracy to 0.31, but produces only 72\% as many draft tokens as the trigram model. This reduction in the quadgram model stems from its longer context. A longer context makes \decode more conservative when generating draft tokens and increases the likelihood of falling back to autoregressive generation. Therefore, while a longer context increases accuracy, it reduces the number of draft tokens, ultimately leading to a 5.1\% slower total decode latency compared to the trigram model.

\section{Related Work} \label{sect:related_work}

{\bf On-device AI.}
Various techniques facilitate the adoption of AI on edge devices. FACIL~\cite{facil} applies processing-in-memory to overcome the memory capacity and bandwidth limitations of edge devices. DecDEC~\cite{decdec} introduces an aggressive low-bit quantization scheme for efficient inference. These works accelerate LLM inference in general, and can be applied to \proposed for further speedup. AppAgent~\cite{appagent} mimics human-like interactions to function without system access. Mobile-Agent~\cite{mobile_agent, mobile_agent_v2} leverages visual capabilities to identify and locate elements in the devices. Overall, these target multi-modal LLM-based agents, whereas \proposed focuses on conventional agentic workflows with text-based LLMs.

{\bf KV cache reuse.}
A rich body of prior work makes use of pre-computed KV caches to reduce runtime latency. Prefix caching~\cite{vllm} reuses KV caches from previous sequences when there are exact prefix matches to the incoming request. However, its reuse is limited when there are mismatches early on in the prompt. Prompt Cache~\cite{prompt_cache} reuses the full KV cache even when there are token matches that do not start at the beginning of the input, and CacheBlend~\cite{cacheblend} selectively recomputes KV caches to maximize reuse while maintaining accuracy. \proposed reuses KV caches by reconstructing the prompts at the text level, targeting the semantic similarity, without any recomputation at runtime.

{\bf Speculative decoding.} \label{page:related_work_specdec}
To facilitate speculative decoding~\cite{sps, specdec}, works like Eagle~\cite{eagle} reduce the training cost of draft LLMs by reusing target model's logits, whereas self-speculative decoding~\cite{draft_and_verify} reuses portions of the target model as draft models, completely removing the need for any retraining. \proposed is distinct from these works in that the lookup table used as the draft model requires no training, and its lightweight, constant time lookup speed allows for speedup directly proportional to the draft token latency. PLD~\cite{pld} constructs an LUT from the user prompt to accelerate input-grounded tasks like summarization. \proposed is different from PLD in that it pinpoints specific parts of the prompt to construct an LUT, and applies selective decoding to avoid the multi-token tax when possible.

\section{Conclusion} \label{sect:conclusion}

We propose \proposed, an on-device agent acceleration solution with no task accuracy degradation. With its two components \prefill and \decode, it accelerates prefill stage of agentic LLMs by 1.97$\times$ and decode stage by 1.73$\times$. Overall, \proposed delivers an end-to-end speedup of 1.61$\times$ on real on-device agents. Because \proposed is a purely software-based solution, it can be applied seamlessly to existing on-device agentic systems. To the best of our knowledge, this is the first work to directly tackle the LLM bottlenecks of on-device agents by leveraging their unique task-level characteristics, under resource-constrained on-device environments.
\begin{acks}
\label{sect:ack}
This work was partly supported by Institute of Information \& Communications Technology Planning \& Evaluation(IITP) grant funded by the Korea government(MSIT) 
(No.RS-2024-00395134, DPU-Centric Datacenter Architecture for Next-Generation AI Devices),
(No.RS-2024-00438851, (SW Starlab) High-performance Privacy-preserving Machine Learning System and System Software), 
(No. RS-2024-00457882, AI Research Hub Project), 
(No.RS-2025-02214652, Development of SoC Technology for AI Semiconductor-Converged Pooled Storage/Memory),
and 
Samsung Electronics Co., Ltd(IO251210-14212-01).
Minsoo Rhu is the corresponding author.
\end{acks}

\bibliographystyle{ACM-Reference-Format}
\bibliography{refs}

\end{document}